\documentclass[conference]{IEEEtran}
\IEEEoverridecommandlockouts
\usepackage{cite}
\usepackage{amsmath,amssymb,amsfonts}
\usepackage{algorithmic}
\usepackage{graphicx}
\usepackage{textcomp}
\usepackage{xcolor}
\usepackage{bm,booktabs,multirow,color}
\usepackage{array}
\usepackage{threeparttable}
\def\BibTeX{{\rm B\kern-.05em{\sc i\kern-.025em b}\kern-.08em
    T\kern-.1667em\lower.7ex\hbox{E}\kern-.125emX}}

\makeatletter
\newcommand{\linebreakand}{%
\end{@IEEEauthorhalign}
\hfill\mbox{}\par
\mbox{}\hfill\begin{@IEEEauthorhalign}
}
\makeatother

\begin{document}

\title{Weakly-supervised anomaly detection for multimodal data distributions
}

\author{\IEEEauthorblockN{1\textsuperscript{st} Xu Tan* \thanks{* Corresponding author}}
\IEEEauthorblockA{\textit{School of Marine Science and Technology} \\
\textit{Northwestern Polytechnical University}\\
Xi'an, China \\
xutan@ieee.org}
\and
\IEEEauthorblockN{2\textsuperscript{nd} Junqi Chen}
\IEEEauthorblockA{\textit{School of Marine Science and Technology} \\
\textit{Northwestern Polytechnical University}\\
Xi'an, China \\
jqchen@ieee.org}
\linebreakand
\IEEEauthorblockN{3\textsuperscript{rd} Sylwan Rahardja}
\IEEEauthorblockA{\textit{School of Computing} \\
	\textit{University of Eastern Finland}\\
	Kuopio, Finland \\
	sylwanrahardja@ieee.org}
\and
\IEEEauthorblockN{4\textsuperscript{th} Jiawei Yang}
\IEEEauthorblockA{\textit{Department of Computing} \\
	\textit{University of Turku}\\
	Turku, Finland \\
	jiaweiyang@ieee.org}
\and
\IEEEauthorblockN{3\textsuperscript{th} Susanto Rahardja*}
\IEEEauthorblockA{\textit{School of Marine Science and Technology} \\
	\textit{Northwestern Polytechnical University}\\
	Xi'an, China}
\IEEEauthorblockA{\textit{Engineering Cluster} \\
	\textit{Singapore Institute of Technology}\\
	Singapore \\
	susantorahardja@ieee.org}
	}

\maketitle

\begin{abstract}
Weakly-supervised anomaly detection can outperform existing unsupervised methods with the assistance of a very small number of labeled anomalies, which attracts increasing attention from researchers. However, existing weakly-supervised anomaly detection methods are limited as these methods do not factor in the multimodel nature of the real-world data distribution. To mitigate this, we propose the Weakly-supervised Variational-mixture-model-based Anomaly Detector (WVAD). WVAD excels in multimodal datasets. It consists of two components: a deep variational mixture model, and an anomaly score estimator. The deep variational mixture model captures various features of the data from different clusters, then these features are delivered to the anomaly score estimator to assess the anomaly levels. Experimental results on three real-world datasets demonstrate WVAD's superiority.
\end{abstract}


\section{Introduction}

Anomaly detection (AD), also known as outlier detection, plays a crucial role in modern data science \cite{yang2023smoothing, yang2024regional}. 
It aims to identify instances that exhibit distinctive characteristics as compared to the majority of the database \cite{tan2022sparse}. 
Conventional AD methods primarily operate in an unsupervised manner, as label acquisition is often difficult and expensive for many applications \cite{yang2021mean, yang2022mipo}. However, unsupervised AD methods are prone to producing false alarms or omissions because they essentially lack prior knowledge \cite{xu2023rosas}.

In practical application scenarios, it is common to have access to a small number of established anomalies, such as patients with a specific diagnosis or detected instances of hacking.
This valuable information is disregarded by unsupervised AD methods. To address this issue, some researchers investigated the use of limited label information for AD, producing weakly-supervised AD \cite{jiang2023weakly}. Generally, weakly-supervised AD methods leverage upon a combination of a small fraction of labeled anomalies and the remaining unlabeled data instances to train a model.
Even with a very small number of labeled anomalies, weakly-supervised AD methods can bring significant performance improvement compared to unsupervised AD methods \cite{jiang2023weakly}, showing its potential in practical usage.

However, existing weakly-supervised AD methods overlooked the multimodal nature of data distributions in many real-world scenarios, where databases contain multiple clusters of data instances, each with distinct characteristics and quantities. These methods simply categorized data instances into normal or anomalous data without considering that instances may originate from different clusters. Consequently, the aforementioned models were trained to capture the global features of the data, failing to identify and capitalize on the local features specific to each cluster. This limitation can lead to the inability to detect local anomalies, such as those lying between two closely positioned clusters.


To address this issue, we propose the Weakly-supervised Variational-mixture-model-based Anomaly Detector (WVAD). WVAD consists of two components: a deep variational mixture model and an anomaly score estimator. The deep variational mixture model aims to extract the latent features of data and categorize them into different clusters based on the latent distributions. Subsequently, several features of an instance are extracted or computed from the mixture model and combined to form a new feature vector. Finally, these new feature vectors are delivered to the anomaly score estimator, which computes the final anomaly scores. Experimental results on three public datasets demonstrate that WVAD outperforms existing state-of-the-art (SOTA) weakly-supervised AD methods.


\section{Related Work}

Recent research on weakly-supervised AD methods for tabular data can be categorized into two groups: the anomaly feature learning and anomaly score learning. Anomaly feature learning methods aim to extract discriminative features between normal and abnormal data using label information. 
Deep Semi-supervised Anomaly Detection (DSAD) \cite{ruff2019deep} minimizes the distance between latent features of unlabeled data and the hypersphere center while maximizing the distance for labeled anomalies. End-to-end Semi-supervised Anomaly Detection (ESAD) \cite{huang2020esad} uses an encoder-decoder-encoder architecture, minimizing the reconstruction errors and l2-norm of latent features for unlabeled data while maximizing them for labeled anomalies. REPEN \cite{pang2018learning} uses contrastive learning to learn discriminative features from pseudo-labeled data and true anomalies. Anomaly-Aware Bidirectional Generative Adversarial Network (AA-BiGAN) \cite{tian2022anomaly} trains the generator and discriminator with different targets based on data labels to make the generated distribution approach the true normal distribution and bypass known anomalies.


Anomaly score learning methods focus on learning an estimator that directly outputs anomaly scores.
Barbora et al. \cite{micenkova2014learning} and Zhao et al. \cite{zhao2018xgbod} proposed using multiple unsupervised anomaly detectors to generate anomaly scores, then combining these scores as a new feature to train a supervised classifier with known labels.
Deviation Networks (DevNet) \cite{pang2019deep} assumes anomaly scores follow a normal distribution. It trains an estimating network that produces scores around the mean for unlabeled data and scores at the right tail for labeled anomalies.
Pairwise Relation prediction Network (PReNet) \cite{pang2023deep} employs contrastive learning by creating three pairwise sets based on data labels and training a network to map data pairs in different sets to different targets.


In general, anomaly feature learning methods excel in generalization ability, while anomaly score learning methods demonstrate superior accuracy on seen data.
To leverage the both advantages, some researchers have explored combining anomaly feature learning and score learning.
Feature Encoding With Autoencoders for Weakly-supervised Anomaly Detection (FEAWAD) \cite{zhou2021feature} consists of a feature autoencoder for data reconstruction and an anomaly score generator. The latent features and reconstruction residues are used by the score generator to estimate labels.
Latent Enhanced classification Deep Generative Model (LEDGM) \cite{xie2021semisupervised} models the latent distributions of normal and abnormal data using a deep mixture variational autoencoder with the help of known anomalies. A classification network classifies data to each distribution and serves as the anomaly detector.
Robust deep Semi-supervised Anomaly Scoring method (RoSAS) \cite{xu2023rosas} utilizes data augmentation by creating new instances and labels via mass interpolation based on known anomalies. It trains a feature representation network to enlarge the distances between the anomaly features and unlabeled data features, and a scoring network to map features to labels.




\section{Methodology}

\subsection{Problem statement}

Given an unlabeled dataset $\mathbf{X}_{\mathrm{U}} \in \mathbb{R}^{N \times D}$ and a labeled anomaly dataset $\mathbf{X}_{\mathrm{A}} \in \mathbb{R}^{M \times D}$, where $N$ and $M$ denote the number of the unlabeled instances and labeled anomalies, respectively; $D$ denotes the dimension of the data. Assuming there are $N_{\mathrm{A}}$ unlabeled anomalies in $\mathbf{X}_{\mathrm{U}}$; $M$ satisfies $M \ll N_{\mathrm{A}}$, as the labeled anomalies only account for a small portion of all anomalies. The final training set $\mathbf{X} = \mathbf{X}_{\mathrm{U}} \cup \mathbf{X}_{\mathrm{A}}$. The target is to measure the anomaly level of each data in $\mathbf{X}_{\mathrm{U}}$.

\subsection{Model architecture}

The proposed WVAD is a combination of the anomaly feature learning and anomaly score learning. It consists of two components: a deep variational mixture model and an anomaly score estimator. The deep variational mixture model is a special variational autoencoder originally designed for clustering. The latent variables are categorized into different clusters according to their dependent probabilities in different prior distributions. There are multiple possible architectures for this model; in this work, the variational deep embedding \cite{jiang2016variational} is used as the backbone model due to its ease of use and performance stability.

Supposing that the data instances in $\mathbf{X}$ come from $K$ different clusters. The objective is to infer the latent variable and cluster label of an input data instance $\mathbf{x}$, and then generate a data distribution in which the likelihood of $\mathbf{x}$ is as high as possible. The generation process works as follows:
\begin{equation}
	\begin{aligned}p_\theta(\mathbf{x},y,\mathbf{z})&=p_\theta(\mathbf{x}|\mathbf{z})p_\theta(\mathbf{z}|y)p(y),\\
		p(y)&=\operatorname{Cat}(\pi),\\
		p(\mathbf{z}|y)&=\mathcal{N}\big(\mathbf{z}\big|\mu_{\mathbf{z},y},\sigma_{\mathbf{z},y}^2\big),
	\end{aligned}
	\label{eq1}
\end{equation}
where $y$ is a discrete latent variable representing the cluster label, and $\mathbf{z} \in \mathbb{R}^{L}$ is a $L$-dimensional continuous variable that represents the latent feature of $\mathbf{x}$. The prior distribution of $y$ is assumed to follow a categorical distribution, while the prior distribution of $\mathbf{z}$ is assumed to be an isotropic Gaussian distribution that is conditioned on $y$. The settings of the parameters of the two distributions will be discussed in Sec. \ref{ssec:ts}.

The target of the inference process is to approximate the true posterior distribution $p(y,\mathbf{z}|\mathbf{x})$, which can be formulated as:
\begin{equation}
	\begin{aligned}
		q_\phi(y,\mathbf{z}|\mathbf{x})&=q(y|\mathbf{x})q_\phi(\mathbf{z}|\mathbf{x}),\\
		q(y|\mathbf{x})=p(y|\mathbf{z})&=\frac{p(y|\boldsymbol{\pi})p(\mathbf{z}|\mu_{\mathbf{z},y},\sigma_{\mathbf{z},y}^2)}{\sum_{y^{\prime}=1}^Kp(y^{\prime}|\boldsymbol{\pi})p(\mathbf{z}|\mu_{\mathbf{z},y^{\prime}},\sigma_{\mathbf{z},y^{\prime}}^2)}, \\
		q_\phi(\mathbf{z}|\mathbf{x})&=\mathcal{N}\big(\mathbf{z}|\mu_\mathbf{z}(\mathbf{x}),\sigma_\mathbf{z}^2(\mathbf{x})\big),
	\end{aligned}
	\label{eq2}
\end{equation}
where $q_\phi(y,\mathbf{z}|\mathbf{x})$ serves as the approximation of $p(y,\mathbf{z}|\mathbf{x})$ and can be factorized into $q(y|\mathbf{x})$ and $q_\phi(\mathbf{z}|\mathbf{x},y)$. $q(y|\mathbf{x})$ is computed in a closed form, while $q_\phi(\mathbf{z}|\mathbf{x},y)$ is parameterized by a neural network.

For unlabeled data, the object of the model is to maximize their likelihood, which can be reformulated as maximizing the evidence lower bound (ELBO) using Jensen’s inequality.
The ELBO can be factorized as:
\begin{equation}
	\begin{aligned}
		&\mathcal{L}_{\mathrm{ELBO}}(\mathbf{x}_{\mathrm{U}};\theta,\phi) \\ 
		&=\mathbb{E}_{q_\phi(y,\mathbf{z}|\mathbf{x}_{\mathrm{U}})}\big[\mathrm{log}p_\theta(\mathbf{x}_{\mathrm{U}}|y,\mathbf{z})+\mathrm{log}p(y)-\mathrm{log}q_\phi(y|\mathbf{x}_{\mathrm{U}}) \\ & \qquad \qquad \qquad \qquad \qquad+\mathrm{log}p(\mathbf{z}|y)-\mathrm{log}q_\phi(y,\mathbf{z}|\mathbf{x}_{\mathrm{U}})]\\
		&=\mathbb{E}_{q_\phi(y,\mathbf{z}|\mathbf{x}_{\mathrm{U}})}[\mathrm{log}p_\theta(\mathbf{x}_{\mathrm{U}}|y,\mathbf{z})]-\mathrm{KL}[q_\phi(y|\mathbf{x}_{\mathrm{U}})||p(y)] \\ & \qquad \qquad \qquad \qquad \qquad \quad \  \ -\mathrm{KL}[q_\phi(y,\mathbf{z}|\mathbf{x}_{\mathrm{U}})||p(\mathbf{z}|y)],
	\end{aligned}
\end{equation}
where $KL[\cdot]$ denotes the Kullback-Leibler divergence. Furthermore, the optimization of the continuous variable $\mathbf{z}$ is carried out using the Stochastic Gradient Variational Bayes estimator and the \textit{reparameterization} trick \cite{kingma2013auto}. Meanwhile, the discrete variable $y$ is enumerated over all possible values to calculate each marginal distribution, which is then summed up together.

The models is not expected to capture the pattern for labeled anomalies. Thus the objective function for them is as follows:
\begin{equation}
	\begin{aligned}
		&\mathcal{L}_{\mathrm{ELBO}}(\mathbf{x}_{\mathrm{A}};\theta,\phi) \\ 
		&=\mathbb{E}_{q_\phi(y,\mathbf{z}|\mathbf{x}_{\mathrm{A}})}[\mathrm{log}p_\theta(\mathbf{x}_{\mathrm{A}}|y,\mathbf{z})]^{-1}-\mathrm{KL}[q_\phi(y|\mathbf{x}_{\mathrm{A}})||p(y)] \\ & \qquad \qquad \qquad \qquad \qquad \quad \  \ -\mathrm{KL}[q_\phi(y,\mathbf{z}|\mathbf{x}_{\mathrm{A}})||p_\theta(\mathbf{z}|y)]^{-1}.
	\end{aligned}
\end{equation}

Once the deep variational mixture model is built, various meaningful features can be extracted or computed from this model. In this work, five distinctive features are identified: the latent cluster probability vector $\mathbf{y}$, the latent feature vector $\mathbf{z}$, the cluster entropy $f_{\mathrm{e}}$, the relative reconstruction error $f_{\mathrm{r}}$, and the cosine similarity $f_{\mathrm{c}}$.

\begin{itemize}
	\item The latent cluster probability vector $\mathbf{y}$ is a $K$-dimensional vector, where the $k$-th element denotes the probability $q(y=k|\mathbf{x})$ as calculated in Eq. \ref{eq2}.
	\item The latent feature vector $\mathbf{z}$ is the reparameterization result from the distribution $\mathcal{N}\big(\mu_\mathbf{z}(\mathbf{x}),\sigma_\mathbf{z}^2(\mathbf{x})\big)$ in the training phase.
	\item The cluster entropy $f_{\mathrm{e}}$ is computed based on $\mathbf{y}$ as follows:
	\begin{equation}
		f_{\mathrm{e}} = -\mathbf{y}^\top\log \mathbf{y}.
	\end{equation}
	$f_{\mathrm{e}}$ indicates the certainty of the model for classifying a data instance to a certain cluster. If $f_{\mathrm{e}}$ is high, the data instance is less likely to fit any cluster, thus carrying a higher probability of being an anomaly.
	\item The relative reconstruction error $f_{\mathrm{r}}$ is computed based on $\mathbf{x}$ as follows: 
	\begin{equation}
		f_{\mathrm{r}} = \frac{||\mathbf{x}-\hat{\mathbf{x}}||_{2}^{2}}{||\mathbf{x}||_{2}^{2}},
	\end{equation}
	where $\hat{\mathbf{x}}$ denotes the expectation of the generation distribution $P_\theta(\mathbf{x}|\mathbf{z})$. $f_{\mathrm{r}}$ reflects whether the model captures the latent feature and the generation mechanism of the data accurately. The higher the $f_{\mathrm{r}}$, the more likely that $\mathbf{x}$ is an anomaly.
	\item The cosine similarity $f_{\mathrm{c}}$ is computed based on $\mathbf{x}$ as follows: 
	\begin{equation}
		f_{\mathrm{c}} = \frac{||\mathbf{x}^\top \hat{\mathbf{x}}||_{2}^{2}}{||\mathbf{x}||_{2}||\hat{\mathbf{x}}||_{2}}.
	\end{equation}
	$f_{\mathrm{c}}$ also reflects whether the model captures the latent features and the generation mechanism of the data accurately. The lower the $f_{\mathrm{c}}$, the more likely that $\mathbf{x}$ is an anomaly.
	
\end{itemize}

The five features are concatenated, producing a new $F$-dimensional feature vector $\mathbf{f}=[\mathbf{y};\mathbf{z};f_{\mathrm{e}};f_{\mathrm{r}};f_{\mathrm{c}}]$, and delivered to the anomaly score estimator. The anomaly score estimator is a fully-connected neural network, and its final layer outputs a single value $s$. $s$ is scaled between 0 and 1 by the sigmoid function. It serves as the estimated anomaly score for $\mathbf{x}$.

\begin{figure}[t]
	\centering
	\includegraphics[scale=1.225]{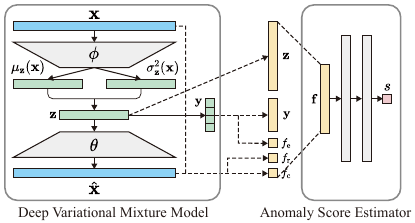}
	\caption{An overview of the whole model architecture of WVAD.}
	\label{fig1}
\end{figure}

Finally, the estimator is optimized by the mini-batch gradient descent algorithm, with the following cross-entropy loss function:
\begin{equation}
	\mathcal{L}_{\mathrm{CE}} = -[\mathbf{t}^\top \log \mathbf{s} + (1-\mathbf{t})^\top\log(1-\mathbf{s})] / B,
\end{equation}
where $B$ denotes the number of data instances in a batch; $\mathbf{s}$ denotes the vector of estimated anomaly scores for the data in a batch; $\mathbf{t}$ denotes the vector of labels for these data, and the label is 1 for labeled anomalies, 0 for unlabeled data (and labeled normal data, if any). An overview of the whole model architecture is shown in Fig. \ref{fig1}.

\subsection{Training strategy}
\label{ssec:ts}

To make the training result more stable, the deep variational mixture model needs to be pretrained.
The deep variational mixture model is treated as a vanilla autoencoder \cite{jiang2016variational}, discarding the variance estimation and reparameterization of $\mathbf{z}$, and pretrained with the following loss function: 
\begin{equation}
	\mathcal{L}_{\mathrm{1}} = \textstyle\sum\big(||\mathbf{x}_{\mathrm{U}}-\hat{\mathbf{x}_{\mathrm{U}}}||_{2}^{2}\big)/n + \textstyle\sum\big(||\mathbf{x}_{\mathrm{A}}-\hat{\mathbf{x}_{\mathrm{A}}}||_{2}^{-2}\big)/m,
\end{equation}
where $n$ and $m$ denote the number of unlabeled data and labeled anomalies in the training batch, respectively.
After training for $e_{1}$ epochs, the expectation-maximization algorithm is applied to the pretrained $\mathbf{z}$ of all data to estimate the parameters of a Gaussian mixture model $\{\pi,\mu_{\mathbf{z},y},\sigma_{\mathbf{z},y}^2\}$. These parameters serve as the initial values of the parameters with the same symbols in Eq. \ref{eq1}.

Next, the deep variational mixture model and anomaly score estimator are combined and trained jointly, using the following joint loss function:
\begin{equation}
	\mathcal{L}_{\mathrm{2}} = -\mathcal{L}_{\mathrm{ELBO}} + \lambda \mathcal{L}_{\mathrm{CE}},
\end{equation}
where $\lambda$ is a weighting parameter, which is set to 0.01 for the first $e_{2}$ epochs, then changes to 1 for the remaining $e_{3}$ epochs. This means the models will focus more on capturing the features of data at the beginning, then divert to training the anomaly score estimator.

\section{Experiments}

\subsection{Datasets}
We used three public datasets from DAMI\footnote{DAMI: www.dbs.ifi.lmu.de/research/outlier-evaluation/DAMI} and ODDS\footnote{ODDS: odds.cs.stonybrook.edu} repositories: \textit{Letter}, \textit{Ionosphere}, and \textit{Satellite}. \textit{Letter} contains artificial anomalies, while \textit{Ionosphere} and \textit{Satellite} contain original anomalies in real-world environments. All contain multimodal data distributions. Their details are summarized as follows:

\subsubsection{Letter}

This dataset contains data of black-and-white rectangular pixel displays for different English capital letters. The data of three letters form the normal classes. For anomalies, a few instances of letters that are not in the normal classes were selected and concatenated with instances from the normal classes, which makes them harder to detect. In total, the dataset contains 1,500 normal data points and 100 anomalies in 32 dimensions.

\subsubsection{Ionosphere}
This dataset contains two kinds of radar data. The good radars, which show evidence of different types of structure in the ionosphere, serve as the normal data. The bad radars, for which signals pass through the ionosphere without being reflected, serve as the anomalies. In total, the dataset contains 225 normal data points and 126 anomalies in 32 dimensions.

\subsubsection{Satellite}
This dataset contains data of multi-spectral satellite images for six types of land. Three types form the normal classes, while the remainder form the anomaly classes. In total, the dataset contains 4,399 normal data and 2,036 anomalies in 36 dimensions. 

\begin{figure}[htbp]
	\centering
	\includegraphics[scale=0.225]{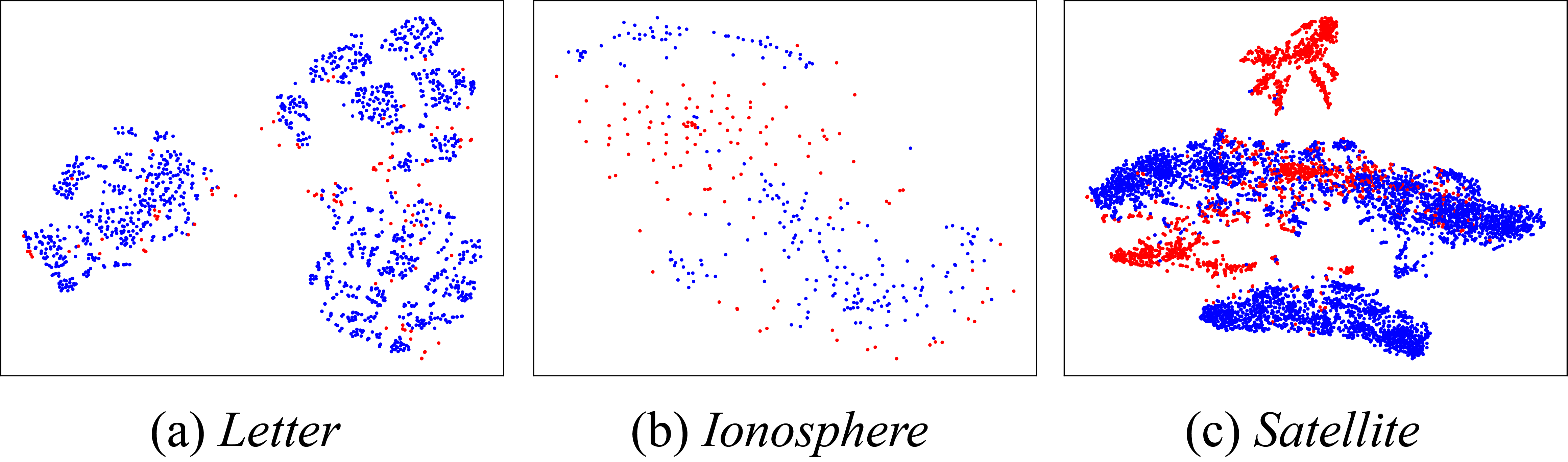}
	\caption{The 2-D visualizations of the \textit{Letter}, \textit{Ionosphere}, and \textit{Satellite} datasets. The normal data are marked in blue, while the anomalies are marked as red. One can observe three obvious data clusters in \textit{Letter}, two clusters in \textit{Ionosphere}, and three clusters in \textit{Satellite}.}
	\label{fig2}
\end{figure}

Fig. \ref{fig2} visualizes these datasets after reducing their dimensions to two dimensions using t-distributed Stochastic Neighbor Embedding (t-SNE)\cite{van2008visualizing}. Before training, all data were standardized using the mean and standard deviation of the dataset. Then, we separately selected 10\% and 5\% of the anomalies as the labeled anomalies for two situations of the weakly-supervised settings.

\subsection{Baselines and settings}
We used six SOTA weakly-supervised AD methods from recent years as our baselines. They are DSAD \cite{ruff2019deep}, DevNet \cite{pang2019deep}, FEAWAD \cite{zhou2021feature}, LEDGM \cite{xie2021semisupervised}, PReNet \cite{pang2023deep}, and RoSAS \cite{xu2023rosas}. Since all these methods, including the proposed WVAD, are neural-network-based methods, we used the same number of network layers and units in each layer for a fair comparison. Specifically, for methods that use an encoder, the dimensions were set to [$D$, $\lfloor D/2 \rfloor$, $\lfloor D/4 \rfloor$, $\lfloor D/8 \rfloor$]; for methods that use a decoder, the dimensions were set to [$\lfloor D/8 \rfloor$, $\lfloor D/4 \rfloor$, $\lfloor D/2 \rfloor$, $D$]; and for methods that use a score estimator, the dimensions were set to [$F$, $F\times$2, $F\times$2, 1]. Other hyperparameters of the baselines were set to the values recommended by their original papers. For the proposed WVAD, the number of data clusters $K$ was set according to the dataset descriptions and visualizations. We set $K$ to 3 for \textit{Letter}, 2 for \textit{Ionosphere}, and 3 for \textit{Satellite}.

As for training, all networks were optimized using the Adam \cite{kingma2014adam} optimizer. The batch size was set to $\lfloor (N+M)/10 \rfloor$, and the number of training epochs was set to 500. For WVAD, \{$e_{1}$, $e_{2}$, $e_{3}$\} were set to \{50, 100, 400\}. Since the number of unlabeled data and labeled anomalies were highly unbalanced, we employed balanced sampling for each batch during training, which resulted in approximately similar numbers for each type of data. Finally, we used the Area Under the Receiver Operating characteristic Curve (AUROC) and the Area Under the Precision-Recall Curve (AUPRC) as our performance metrics. Both range from 0 to 1, where 1 indicates the best performance.

\subsection{Experimental results}
\begin{figure}[htbp]
	\centering
	\includegraphics[scale=0.26]{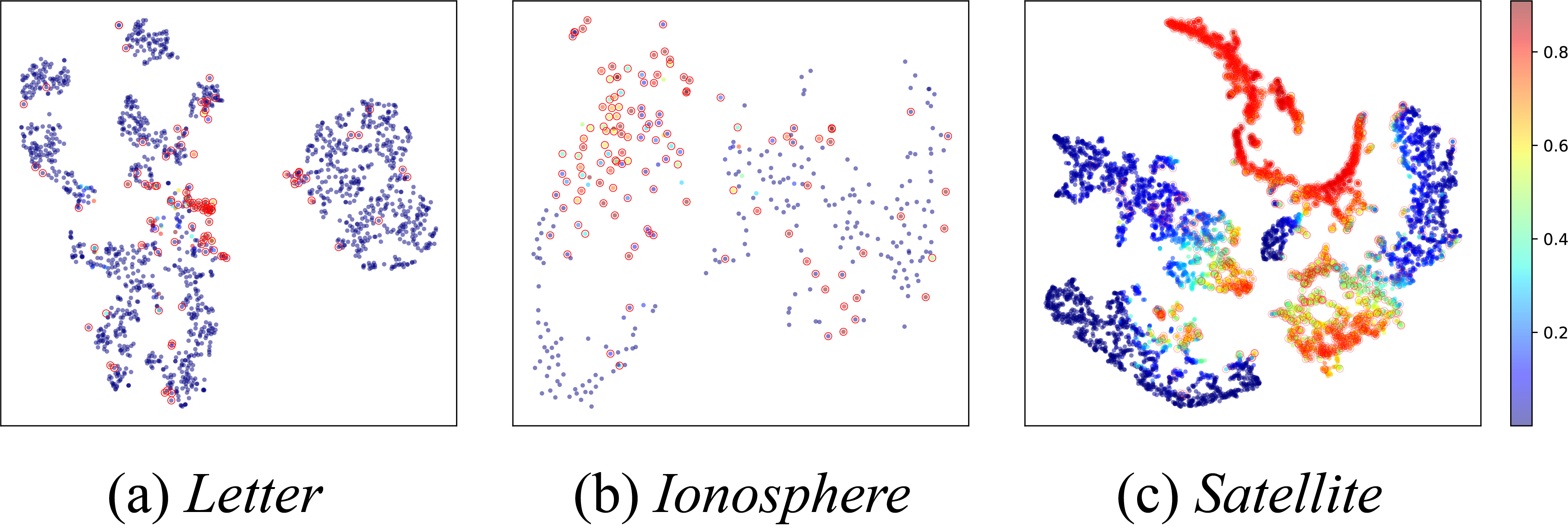}
	\caption{The 2-D visualizations of the latent features $\mathbf{z}$ extracted from the WVAD, colored by the estimated anomaly scores. The warmer the color, the higher the anomaly score. The true anomalies are surrounded by red circles.}
	\label{fig3}
\end{figure}

Firstly, we evaluated the proposed WVAD on the three datasets with a label ratio of 10\%. We used t-SNE to visualize the latent features $\mathbf{z}$ extracted from the model and colored them according to the estimated anomaly scores. As shown in Fig. \ref{fig3}, the warmer the color, the higher the anomaly score. The true anomalies were surrounded by red circles. It was observed that in \textit{Letter} and \textit{Ionosphere}, the anomalies between different clusters could be detected very well. \textit{Ionosphere} was unique since there were clusters being dominated by anomalies. WVAD could still detect these anomalies well because of the latent cluster probability vector $\mathbf{y}$ it used. If one data cluster contained too many labeled anomalies, then instances with similar cluster probabilities may also be anomalies.

\begin{table}[ht]
	\centering
	\caption{Performance comparisons of all methods under a label ratio of 10\%}
	\label{tab1}
	\scalebox{0.84}{
	\begin{tabular}{c|cc|cc|cc}
		\toprule
		\multirow{2}{*}{Methods} & \multicolumn{2}{c|}{Letter}     & \multicolumn{2}{c|}{Ionosphere} & \multicolumn{2}{c}{Satellite}   \\
		& AUROC          & AUPRC          & AUROC          & AUPRC          & AUROC          & AUPRC          \\ \midrule
		DSAD \cite{ruff2019deep}                    & 0.827          & 0.459          & 0.775          & 0.649          & 0.781          & 0.701          \\
		DevNet \cite{pang2019deep}                  & 0.744          & 0.340          & 0.819          & 0.792          & 0.824          & 0.771          \\
		FEAWAD \cite{zhou2021feature}                  & \textbf{0.888} & 0.516          & 0.783          & 0.756          & 0.888          & 0.848          \\
		LEDGM \cite{xie2021semisupervised}                   & 0.614          & 0.142          & 0.820          & 0.747          & 0.838          & 0.786          \\
		PReNet \cite{pang2023deep}                  & 0.625          & 0.280          & 0.551          & 0.499          & 0.804          & 0.780          \\
		RoSAS \cite{xu2023rosas}                   & 0.735          & 0.407          & 0.518          & 0.501          & 0.880          & 0.828          \\
		WVAD*                     & 0.882          & \textbf{0.565} & \textbf{0.925} & \textbf{0.870} & \textbf{0.934} & \textbf{0.883} \\ \bottomrule
	\end{tabular}}
	\begin{tablenotes}
		\footnotesize
		\item * Proposed method.
	\end{tablenotes}
\end{table}

\begin{table}[ht]
	\centering
	\caption{Performance comparisons of all methods under a label ratio of 5\%}
	\label{tab2}
	\scalebox{0.84}{
	\begin{tabular}{c|cc|cc|cc}
		\toprule
		\multirow{2}{*}{Methods} & \multicolumn{2}{c|}{Letter}     & \multicolumn{2}{c|}{Ionosphere} & \multicolumn{2}{c}{Satellite}   \\
		& AUROC          & AUPRC          & AUROC          & AUPRC          & AUROC          & AUPRC          \\ \midrule
		DSAD \cite{ruff2019deep}                     & 0.799          & 0.353          & 0.695          & 0.543          & 0.784          & 0.661          \\
		DevNet \cite{pang2019deep}                   & 0.729          & 0.246          & 0.705          & 0.674          & 0.811          & 0.762          \\
		FEAWAD \cite{zhou2021feature}                   & 0.841          & 0.391          & 0.756          & 0.718          & 0.852          & 0.816          \\
		LEDGM \cite{xie2021semisupervised}                    & 0.621          & 0.117          & 0.502          & 0.476          & 0.854          & 0.807          \\
		PReNet \cite{pang2023deep}                   & 0.560          & 0.177          & 0.513          & 0.460          & 0.703          & 0.672          \\
		RoSAS \cite{xu2023rosas}                    & 0.648          & 0.278          & 0.474          & 0.429          & 0.824          & 0.741          \\
		WVAD*                     & \textbf{0.852} & \textbf{0.478} & \textbf{0.850} & \textbf{0.748} & \textbf{0.905} & \textbf{0.836} \\ \bottomrule
	\end{tabular}}
	\begin{tablenotes}
		\footnotesize
		\item * Proposed method.
	\end{tablenotes}
\end{table}

Secondly, we compared the performance of WVAD with all baselines on the three datasets under both label ratio conditions. All methods were run 10 times with different random seeds, and the average performance was collected. The results are shown in Tab. \ref{tab1} and Tab. \ref{tab2}. It was observed that under a label ratio of 10\%, WVAD surpassed all the baselines under both metrics on all three datasets, except for AUROC on \textit{Letter}, for which FEAWAD was slightly better than WVAD. When the label ratio was set to 5\%, WVAD significantly outperformed all the baselines. The results strongly supported the superiority of the proposed WVAD.

\section{Conclusions}
In this paper, we proposed WVAD, a weakly-supervised anomaly detector designed for multimodal data distributions. WVAD first models the data distributions using the deep variational mixture model, then extracts five distinct features to estimate the final anomaly scores using the anomaly score estimator. The experimental results showed that WVAD could detect different types of anomalies and surpassed six SOTA weakly-supervised AD methods. In the future, WVAD can be further improved by employing a better deep mixture model or elaborating on the features for specific applications.

\bibliographystyle{IEEEtran}
\bibliography{refs}

\end{document}